\documentclass[letterpaper]{article} 
\usepackage{aaai2027}  
\usepackage[hyphens]{url}  
\usepackage{graphicx} 
\usepackage{natbib}  
\usepackage{caption} 
\usepackage{algorithm}
\usepackage{algorithmic}
\usepackage{amssymb}

\usepackage{amsmath,amsfonts,bm}

\def\figref#1{figure~\ref{#1}}
\def\Figref#1{Figure~\ref{#1}}

\def\eqref#1{equation~\ref{#1}}

\def\1{\bm{1}}

\def\rvepsilon{{\mathbf{\epsilon}}}

\def\vzero{{\bm{0}}}

\def\vtheta{{\bm{\theta}}}

\def\vc{{\bm{c}}}
\def\vd{{\bm{d}}}
\def\ve{{\bm{e}}}

\def\vr{{\bm{r}}}
\def\vs{{\bm{s}}}

\def\vv{{\bm{v}}}

\def\vx{{\bm{x}}}

\def\vz{{\bm{z}}}

\def\mI{{\bm{I}}}

\def\mU{{\bm{U}}}

\def\mW{{\bm{W}}}

\DeclareMathAlphabet{\mathsfit}{\encodingdefault}{\sfdefault}{m}{sl}
\SetMathAlphabet{\mathsfit}{bold}{\encodingdefault}{\sfdefault}{bx}{n}
\newcommand{\tens}[1]{\bm{\mathsfit{#1}}}

\def\tH{{\tens{H}}}

\def\gB{{\mathcal{B}}}

\def\gD{{\mathcal{D}}}

\def\gF{{\mathcal{F}}}

\def\gN{{\mathcal{N}}}

\def\gS{{\mathcal{S}}}

\def\gU{{\mathcal{U}}}

\def\sR{{\mathbb{R}}}

\newcommand{\E}{\mathbb{E}}
\newcommand{\Ls}{\mathcal{L}}

\def\tabref#1{table~\ref{#1}}
\def\Tabref#1{Table~\ref{#1}}
\def\twotabref#1#2{tables~\ref{#1} and~\ref{#2}}

\def\vphi{{\bm{\phi}}}
\def\vgamma{{\bm{\gamma}}}
\DeclareMathOperator{\Euler}{Euler}
\DeclareMathOperator{\Unfold}{Unfold}
\DeclareMathOperator{\Fold}{Fold}
\DeclareMathOperator{\Mod}{Mod}
\DeclareMathOperator{\Norm}{Norm}
\DeclareMathOperator{\Scale}{scale}
\DeclareMathOperator{\Shift}{shift}
\DeclareMathOperator{\Attn}{Attn}
\DeclareMathOperator{\LPIPS}{LPIPS}
\newcommand{\normone}[1]{\left\lVert #1 \right\rVert_1}
\newcommand{\normtwo}[1]{\left\lVert #1 \right\rVert_2}

\usepackage{xcolor}
\newcommand{\cao}[1]{\textcolor{black}{#1}}
\newcommand{\resultgroup}[1]{\multicolumn{9}{@{}l@{}}{\makebox[0pt][l]{{\color{black!10}\rule[-0.45ex]{449.3pt}{2.1ex}}}\hspace{\tabcolsep}\emph{#1}} \\}
\usepackage{newfloat}
\usepackage{listings}
\DeclareCaptionStyle{ruled}{labelfont=normalfont,labelsep=colon,strut=off} 
\floatstyle{ruled}
\newfloat{listing}{tb}{lst}{}
\floatname{listing}{Listing}

\usepackage{booktabs}

\title{\cao{PixelIR: Fidelity–Perception Decoupling via Pixel-Space Image–Residual Flow Matching for Efficient One-Step Real-World Super-Resolution}}
\author{
  Bingtian Qiao\textsuperscript{\rm 1,\rm 2}\thanks{Work done during an internship at Shanghai Jiao Tong University.},
  Yue Shi\textsuperscript{\rm 1,\rm 3},
  Yong Guo\textsuperscript{\rm 1},
  Wenjun Zhang\textsuperscript{\rm 1},
  Jiezhang Cao\textsuperscript{\rm 1}\corresponding
}
\affiliations{
  \textsuperscript{\rm 1}Shanghai Jiao Tong University
  \textsuperscript{\rm 2}Fuzhou University
  \textsuperscript{\rm 3}Shanghai AI Laboratory\\
  caojiezhang@sjtu.edu.cn
}

\makeatletter
\let\pixelpyrmaketitle\@maketitle
\def\@maketitle{%
  \pixelpyrmaketitle
  \vspace{-13mm}
  \begin{center}
  \makebox[\textwidth][c]{%
    \includegraphics[height=0.325\textwidth]{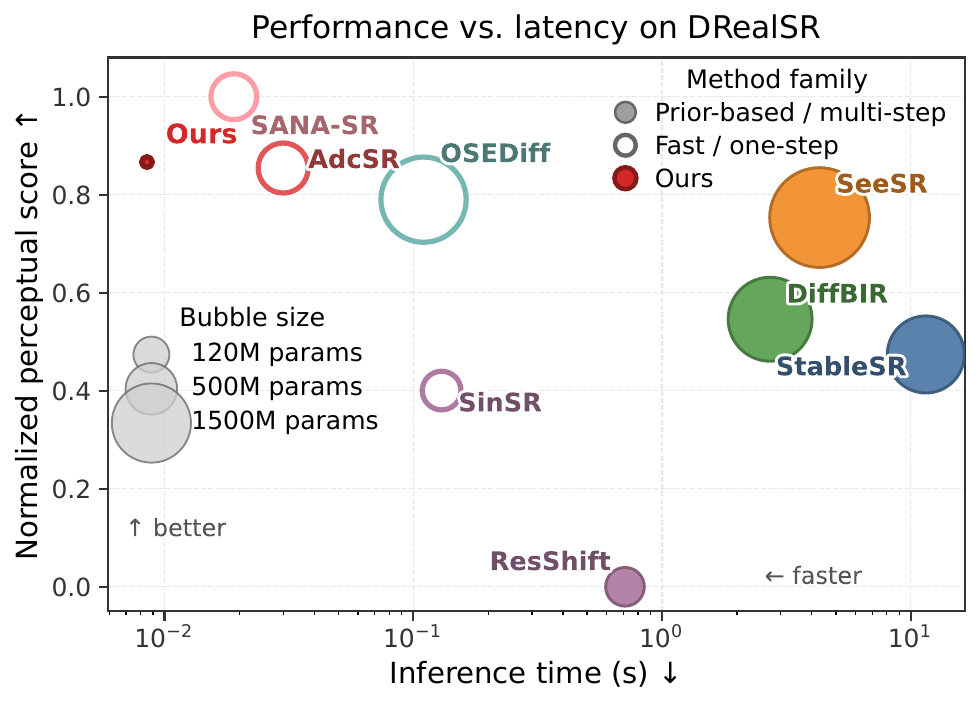}\hspace{0.01\textwidth}
    \includegraphics[height=0.325\textwidth]{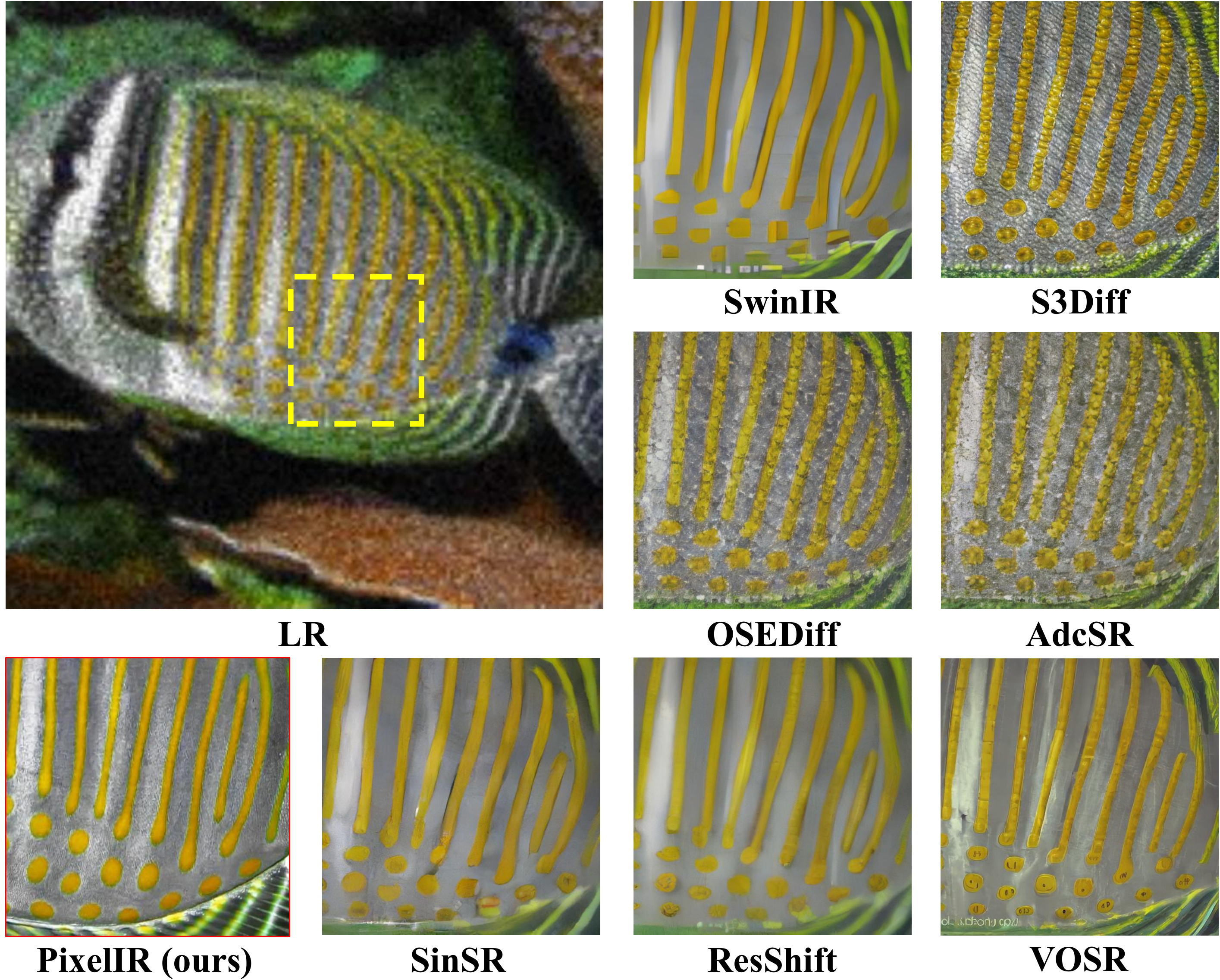}%
  }
  \captionof{figure}{\textbf{The compact PixelIR student achieves a favorable quality--efficiency trade-off for Real-ISR. Left:} DRealSR perceptual score versus latency. \textbf{Right:} comparison against seven baselines.}
  \label{fig:teaser}
  \end{center}
}
\makeatother

\begin{document}

\maketitle

\begin{abstract}
\cao{Real-world image super-resolution (Real-ISR) aims to preserve structures supported by the degraded observation while reconstructing perceptually realistic details.}
However, existing Real-ISR methods largely optimize fidelity and perceptual quality within a \cao{shared} network, \cao{causing the two objectives to interfere throughout training and making their balance difficult to control.}
\cao{Recent one-step methods reduce sampling steps, yet often inherit both this coupled optimization behavior and the expensive high-resolution backbone of their multi-step predecessors.}
We argue that efficient Real-ISR requires not only a shorter sampling trajectory, but also specialized modeling of faithful reconstruction and perceptual detail synthesis. 
\cao{Based on this insight, we propose PixelIR, a fidelity–perception decoupling framework built upon pixel-space image–residual flow matching.}
\cao{PixelIR first learns an image flow that maps the degraded observation to a faithful reconstruction.
Then, a residual flow synthesizes the missing perceptual details from noise without repeatedly relearning or overwriting the complete restoration solution. 
We further distill the teacher into a deployment-oriented one-step student within a coarse-to-fine pyramid architecture. 
Extensive experiments show that PixelIR achieves leading PSNR, SSIM, and LPIPS on both RealSR and DRealSR. The compact student completes pixel-space restoration in a single evaluation with only 32.9M parameters, 89.7G MACs, and 8.5ms latency, demonstrating a strong practical fidelity–perception–efficiency balance.}
\end{abstract}

\section{Introduction}

Real-world image super-resolution (Real-ISR) aims to recover high-quality images from observations degraded by unknown blur, noise, compression, and sensor artifacts. Existing methods face a fundamental fidelity--perception trade-off~\cite{blau2018perception}: pixel-wise objectives favor accurate but over-smoothed conditional means, whereas perceptual and adversarial objectives recover realistic textures at the risk of deviating from the reference. When optimized within one generator, reweighting these objectives typically selects another point on the same empirical Pareto frontier rather than improving both. Their balance therefore remains central to practical Real-ISR.

Recent years have witnessed the growing success of generative restoration methods for Real-ISR. Compared with regression-based approaches, diffusion and flow-based models better address real-world degradations by synthesizing plausible textures and realistic details. However, existing methods still occupy only part of the design space. Multi-step methods repeatedly evaluate large high-resolution models, whereas one-step methods avoid iterative sampling but often retain expensive denoisers and auxiliary encoders. More fundamentally, both families optimize fidelity and perception within a shared backbone, leaving the objectives to compete throughout training. Consequently, neither family simultaneously improves the fidelity--perception operating point and enables lightweight one-step deployment. Current Real-ISR is therefore limited by sampling depth and costly, coupled high-resolution modeling.

This observation motivates us to revisit Real-ISR from two important perspectives: structural decoupling and efficient one-step deployment. To this end, we propose PixelIR, a decouple-then-distill framework. Specifically, we first construct a two-stage teacher, in which the first stage learns a faithful base reconstruction and is then frozen, while the second stage synthesizes the missing residual details conditioned on this base. This design assigns fidelity reconstruction and perceptual generation to separately optimized networks, preventing the perceptual objective from repeatedly renegotiating the structural solution. Compared with coupled single-stage training, the resulting teacher improves PSNR, LPIPS, and DISTS simultaneously. The gain therefore shifts the observed fidelity--perception frontier outward instead of selecting another trade-off point.

The decoupled teacher establishes the desired operating point but remains expensive: its two 4-step networks contain 370.1M parameters and require eight evaluations. We therefore distill its restoration mapping into a purpose-built one-step student. The student progressively increases feature resolution from $32^2$ to $512^2$ while reducing channel width from 256 to 48, concentrating capacity at semantic scales and keeping high-resolution processing narrow. Cellwise compress-and-expand blocks cap attention at 1024 tokens in the three finest stages. Complementary teacher and ground-truth supervision transfers perceptual detail without bounding the student by the teacher's reconstruction error.

Extensive experiments on standard Real-ISR benchmarks demonstrate that PixelIR achieves the best PSNR, SSIM, and LPIPS on both real-capture benchmarks, while its compact student surpasses all prior methods in CLIP-IQA on DIV2K. More importantly, the 32.9M-parameter student performs the complete LR-to-HR mapping in one pixel-space evaluation with 89.7G MACs and 8.5ms latency on an RTX PRO 6000---the lowest parameter count, MACs, and latency among the compared methods. This corresponds to $5.3\times$ fewer parameters than the next-smallest deployment stack and $4.6\times$ fewer MACs than SANA-SR. Without VAE or text branches, the pyramid transfers the improved fidelity--perception operating point to an efficient deployment regime.

Our main contributions are summarized as follows: 
\begin{itemize}
    \item \cao{We identify coupled optimization between faithful reconstruction and perceptual detail synthesis as an important limitation of generative Real-ISR, and propose PixelIR, a decouple-then-distill framework that separately organizes restoration quality and deployment efficiency.}
    \item \cao{We introduce pixel-space image–residual flow matching, in which a frozen image flow establishes a faithful structural anchor and a conditional residual flow performs complementary perceptual refinement. Under matched settings, this decomposition improves PSNR, LPIPS, and DISTS over coupled single-stage optimization.}
    \item \cao{We design a 32.9M-parameter coarse-to-fine student with bounded fine-scale attention and teacher–ground-truth supervision, enabling one-step restoration with 89.7G MACs while retaining leading fidelity and reference-based perception.}
\end{itemize}

\section{Related Work}
\label{sec:related}

\noindent\textbf{Perception--distortion in Real-ISR.}
Earlier SR approaches rely on feed-forward regression with pixel-wise objectives~\cite{dong2016srcnn,shi2016espcn,kim2016vdsr,zhang2018rcan,chen2021ipt,liang2021swinir,chen2023hat}. They favor high PSNR but average plausible high-frequency completions, producing over-smoothed results. Perceptual and adversarial training improves realism~\cite{ledig2017srgan,wang2018esrgan,liang2022ldl}, while BSRGAN~\cite{zhang2021bsrgan} and Real-ESRGAN~\cite{wang2021realesrgan} extend it to unknown real degradations. Optimizing these objectives in one restorer exposes the perception--distortion trade-off~\cite{blau2018perception}: perceptual gains can reduce distortion fidelity.

Generative Real-ISR increasingly adapts powerful generative models~\cite{ho2020ddpm,song2021scoresde,rombach2022ldm,peebles2023dit,podell2024sdxl,esser2024sd3,saharia2023sr3}. StableSR~\cite{wang2024stablesr}, DiffBIR~\cite{lin2024diffbir}, SeeSR~\cite{wu2024seesr}, PASD~\cite{yang2024pasd}, and SUPIR~\cite{yu2024supir} introduce different priors and conditions; ResShift~\cite{yue2023resshift} models the residual shift between low- and high-resolution images; and CCSR~\cite{sun2025ccsr} separates structure and detail across sampling stages. Yet fidelity and perception remain negotiated within a shared generator. PixelIR instead assigns them to separately optimized networks and freezes the fidelity stage, targeting the Pareto frontier rather than only strengthening the prior.

\begin{figure*}[t]
\centering
\includegraphics[width=0.99\textwidth,trim=0 0 0 0,clip]{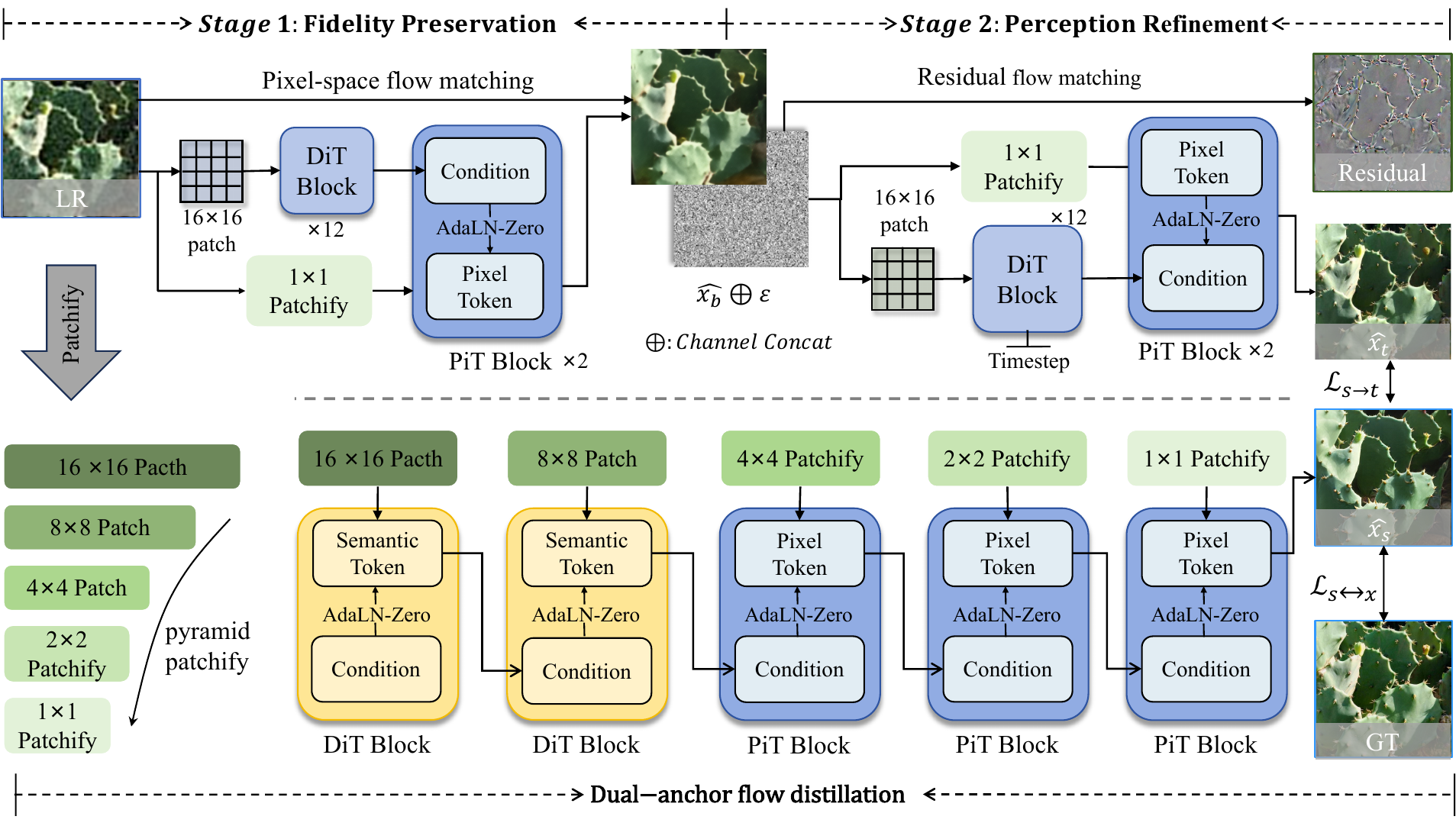}
\caption{\textbf{Overview of PixelIR.} Given an upsampled LQ input, PixelIR first obtains a faithful base reconstruction $\hat{\vx}_{\mathrm b}$ with a 4-step image flow, then generates a complementary residual $\hat{\vr}$ with a 4-step residual flow conditioned on the frozen base and forms the teacher output $\hat{\vx}_{\mathrm t}$ through residual composition. Both stages use a dual-granularity pixel transformer with semantic DiT and pixel-level PiT blocks. The teacher is further distilled into a progressively narrowed five-stage student $\gS_{\vphi}$, guided by the teacher anchor $\Ls_{\mathrm{s}\rightarrow\mathrm{t}}$ and the bidirectional ground-truth anchors $\Ls_{\mathrm{s}\leftrightarrow\mathrm{x}}$ for efficient one-step deployment.}
\label{fig:architecture}
\end{figure*}

\noindent\textbf{Distillation for efficient SR.}
Another thread reduces inference to one or a few network evaluations~\cite{song2021ddim}. Progressive distillation~\cite{salimans2022progressive}, consistency models~\cite{song2023consistency}, and InstaFlow~\cite{liu2024instaflow} compress iterative generation~\cite{yin2024dmd,yin2024dmd2,xu2024ufogen,sauer2024add}. For Real-ISR, SinSR~\cite{wang2024sinsr} distills ResShift with inverse consistency; OSEDiff~\cite{wu2024osediff} adopts variational score distillation; S3Diff~\cite{zhang2024s3diff} introduces degradation guidance; and AdcSR~\cite{chen2025adcsr} combines distillation with structural pruning. These methods compress an existing trajectory, whereas PixelIR first improves the teacher's fidelity--perception operating point and then distills its mapping and deployment cost.

\noindent\textbf{Efficient generative architectures.}
Beyond shortening the sampling trajectory, a complementary line redesigns the backbone for high-resolution efficiency. PixelFlow~\cite{chen2025pixelflow} cascades progressively higher-resolution flow stages, while PixelDiT~\cite{yu2026pixeldit} compresses local pixel groups into bounded token sequences within each block. SANA-SR~\cite{qiao2026sanasr} combines compact latent tokens with linear attention for one-step restoration. PixelIR instead couples progressive resolution with within-block token compression, increasing resolution while reducing channel width and bounding fine-scale attention as distillation preserves the decoupled teacher's operating point. Unlike compact latent-token designs, the student performs the complete restoration directly in pixel space without relying on a latent autoencoder at inference. This retains an explicit pixel-domain restoration path throughout deployment.

\section{Proposed Method}

\noindent\textbf{Preliminary.}
\label{sec:prelim}

Real-ISR admits multiple high-frequency completions for the same degraded observation. We formulate this conditional generation process with rectified flow~\cite{liu2023rectflow,lipman2023flowmatching}, using a common pixel-space parameterization for the two 4-step teacher flows and one-step student. Let $\vc,\vx\in\sR^{3\times H\times W}$ denote the upsampled low-quality observation and its high-quality target. For source $\vs$, destination $\vd$, and $t\in[0,1]$, the interpolation state $\vz_t$ and target velocity $\vv^\star$ are
\begin{equation}
\vz_t = t\,\vd + (1-t)\,\vs, \qquad \vv^\star = \vd - \vs.
\label{eq:flow}
\end{equation}
A velocity network $\gF_{\vtheta}$ is optimized by
\begin{equation}
\Ls_{\mathrm{v}} = \E_{t,\vs,\vd}\left[\normtwo{\gF_{\vtheta}(\vz_t,t)-\vv^\star}^2\right],
\label{eq:lv}
\end{equation}
where $t$ follows a logit-normal distribution, and $\hat{\vd}=\vz_t+(1-t)\gF_{\vtheta}(\vz_t,t)$ estimates the destination at any state. At inference, an $N$-step Euler solver transports $\vz_0=\vs$ toward $\vd$; for $N=1$, $\hat{\vd}=\vs+\gF_{\vtheta}(\vs,0)$. PixelIR applies this formulation directly in RGB space without an auxiliary autoencoder, for both the multi-step teacher and one-step deployment model.

\subsection{Fidelity-Oriented Image Flow}
\label{sec:fidelity}

The first stage models fidelity preservation as an image flow from $\vc$ to $\vx$ by setting $\vs=\vc$ and $\vd=\vx$ in \eqref{eq:flow}. A base restorer $\gB_{\vtheta_{\mathrm b}}$ predicts the corresponding velocity, and a 4-step Euler trajectory gives
\begin{equation}
\hat{\vx}_{\mathrm b}=\Euler_4\big(\gB_{\vtheta_{\mathrm b}};\vc\big).
\label{eq:base}
\end{equation}

For an estimate $\hat{\vx}$, we define $\Ls_1(\hat{\vx})=\normone{\hat{\vx}-\vx}$, $\Ls_{\mathrm p}(\hat{\vx})=\LPIPS(\hat{\vx},\vx)$, and $\Ls_{\mathrm f}(\hat{\vx})$ as the cosine distance between its projected tokens and frozen DINOv2 features~\cite{oquab2024dinov2}. Let $\Ls_q^{\mathrm b}=\Ls_q(\hat{\vx}_{\mathrm b})$ for $q\in\{1,\mathrm p,\mathrm f\}$. The fidelity objective is
\begin{equation}
\Ls_{\mathrm{fid}}=\lambda_{\mathrm v}\Ls_{\mathrm v}^{\mathrm b}+\lambda_1\Ls_1^{\mathrm b}+\lambda_{\mathrm p}\Ls_{\mathrm p}^{\mathrm b}+\lambda_{\mathrm f}\Ls_{\mathrm f}^{\mathrm b}.
\label{eq:lfid}
\end{equation}
Here, $\Ls_{\mathrm v}^{\mathrm b}$ instantiates \eqref{eq:lv}. The base restorer uses the dual-granularity pixel transformer in \figref{fig:architecture}: a 12-block, 768-dimensional semantic DiT operates on $16\times16$ patches, and a 2-block, 64-dimensional PiT performs local pixel refinement. After training, the 184.8M-parameter image flow is frozen as the perceptual stage's fidelity anchor.

\subsection{Perception-Oriented Residual Flow}
\label{sec:perception}

The second stage models detail refinement as a residual flow conditioned on the frozen base, with target
\begin{equation}
\vr=\vx-\hat{\vx}_{\mathrm b}.
\label{eq:residual}
\end{equation}
We set $\vs=\rvepsilon$ and $\vd=\vr$, where $\rvepsilon\sim\gN(\vzero,\mI)$ is a Gaussian random vector. The residual-flow state and velocity are therefore $\vz_t^{\mathrm r}=t\vr+(1-t)\rvepsilon$ and $\vv_{\mathrm r}^{\star}=\vr-\rvepsilon$, respectively. A detail network $\gD_{\vtheta_{\mathrm d}}$ receives the flow state and the frozen base through channel-wise concatenation:
\begin{equation}
\tilde{\vz}_t^{\mathrm r}=\big[\,\vz_t^{\mathrm r};\hat{\vx}_{\mathrm b}\,\big]
\in\sR^{6\times H\times W}.
\label{eq:concat}
\end{equation}
The 4-step residual trajectory starts from $\rvepsilon$ and produces $\hat{\vr}=\Euler_4(\gD_{\vtheta_{\mathrm d}};\rvepsilon,\hat{\vx}_{\mathrm b})$. The final teacher output is
\begin{equation}
\hat{\vx}_{\mathrm t}=\hat{\vx}_{\mathrm b}+\hat{\vr}.
\label{eq:compose}
\end{equation}
Let $\Ls_q^{\mathrm t}=\Ls_q(\hat{\vx}_{\mathrm t})$ for $q\in\{1,\mathrm p,\mathrm f\}$. The perceptual-stage objective is
\begin{equation}
\Ls_{\mathrm{per}}=\lambda_{\mathrm v}\Ls_{\mathrm v}^{\mathrm r}+\lambda_1\Ls_1^{\mathrm t}+\lambda_{\mathrm p}\Ls_{\mathrm p}^{\mathrm t}+\lambda_{\mathrm f}\Ls_{\mathrm f}^{\mathrm t}+\lambda_{\mathrm a}\Ls_{\mathrm{adv}}.
\label{eq:ldet}
\end{equation}
where $\Ls_{\mathrm v}^{\mathrm r}$ matches $\vv_{\mathrm r}^{\star}$, and $\Ls_{\mathrm{adv}}$ is the generator-side relativistic average objective~\cite{wang2018esrgan} using a conditional multi-scale PatchGAN. The detail flow retains the dual-granularity transformer and expands its input projection from three to six channels for \eqref{eq:concat}. It contains 185.3M parameters, giving 370.1M parameters and eight evaluations for the complete teacher. Freezing $\gB_{\vtheta_{\mathrm b}}$ assigns faithful reconstruction and perceptual residual synthesis to separately optimized flows, while residual composition preserves the base reconstruction as the reference solution.

\noindent\textbf{Prior-enhanced teacher.} The PixelIR-L teacher follows the same image--residual flow formulation, base conditioning, 4-step trajectories, and residual composition. It scales the perceptual flow with a pretrained 1.3B text-to-image PixelDiT prior and rank-32 LoRA. The fidelity flow remains frozen, while the prior-initialized residual flow uses the same objective in \eqref{eq:ldet}, yielding a higher-capacity perceptual variant.

\subsection{One-Step Flow Distillation}
\label{sec:distill}

For the lightweight branch, PixelIR compresses the 370.1M-parameter, eight-evaluation teacher into a 32.9M-parameter one-step student $\gS_{\vphi}$; PixelIR-L applies the same distillation formulation with a deeper 132.8M pyramid that preserves the five resolutions and widths. The lightweight student extends the dual-granularity design into a five-stage pyramid whose resolution increases as its channel width decreases, allocating most capacity to semantic reasoning while keeping fine-scale processing lightweight.

\noindent\textbf{Bounded-attention pyramid.} For a feature tensor $\tH\in\sR^{C\times R\times R}$, we partition the spatial grid into non-overlapping $p\times p$ cells and flatten every cell into one token:
\begin{equation}
\Unfold_p(\tH)\in\sR^{(R/p)^2\times p^2C}.
\label{eq:unfold}
\end{equation}
With timestep embedding $\ve_t$ and projections $\mW_{\mathrm c}\in\sR^{p^2C\times d}$ and $\mW_{\mathrm e}\in\sR^{d\times p^2C}$, the block computes
\begin{equation}
\mU=\Unfold_p\big(\Mod(\Norm(\tH);\ve_t)\big)\mW_{\mathrm c}
\in\sR^{(R/p)^2\times d},
\label{eq:compress}
\end{equation}
\begin{equation}
\hat{\mU}=\Attn(\mU)
\in\sR^{(R/p)^2\times d},
\label{eq:attn}
\end{equation}
\begin{equation}
\tH\leftarrow\tH+\vgamma\odot\Fold_p\big(\hat{\mU}\mW_{\mathrm e}\big)
\in\sR^{C\times R\times R},
\label{eq:expand}
\end{equation}
where $\Mod(\tH;\ve_t)=\tH\odot(1+\Scale(\ve_t))+\Shift(\ve_t)$, $\vgamma$ is a timestep gate, $\Fold_p$ inverts $\Unfold_p$, and $\Attn$ denotes rotary self-attention~\cite{su2024rope}. A second residual branch applies a feed-forward network. At $p=1$, the block attends the $R\times R$ grid; at $p=R/32$, it attends $32^2=1024$ tokens.

$\gS_{\vphi}$ contains five stages at $R_i\in\{32,64,128,256,512\}$, connected by learned $2\times$ pixel-shuffle upsampling. As shown in \tabref{tab:student-arch}, width decreases as $256{\to}160{\to}112{\to}72{\to}48$, with most depth at the first semantic stage. The first two stages use $p_i=1$; the three finer stages set $p_i=R_i/32$ and retain 1024 tokens. The input is patchified with $p=16$ and projected to the first-stage width; a final $3\times3$ convolution predicts the three-channel velocity $\vv_{\vphi}(\vz_t,t)$.

\begin{table}[t]\centering\small
\begin{tabular}{cccccc}
\toprule
Stage & $R_i$ & $p_i$ & $C_i$ & $L_i$ & tokens \\
\midrule
0 & 32  & 1  & 256 & 6 & 1024 \\
1 & 64  & 1  & 160 & 3 & 4096 \\
2 & 128 & 4  & 112 & 2 & 1024 \\
3 & 256 & 8  & 72  & 2 & 1024 \\
4 & 512 & 16 & 48  & 2 & 1024 \\
\bottomrule
\end{tabular}
\caption{Student pyramid configuration.}
\vspace{-5mm}
\label{tab:student-arch}
\end{table}

\noindent\textbf{Dual-anchor flow distillation.} The student learns a one-step pixel-space flow from source $\vc$ to target $\vx$:
\begin{equation}
\hat{\vx}_{\mathrm s}=\vc+\vv_{\vphi}(\vc,0).
\label{eq:student-out}
\end{equation}
For the frozen teacher output $\hat{\vx}_{\mathrm t}$, student-to-teacher supervision transfers the teacher restoration behavior:
\begin{equation}
\Ls_{\mathrm{s}\rightarrow\mathrm{t}}=\alpha_1\normone{\hat{\vx}_{\mathrm s}-\hat{\vx}_{\mathrm t}}
+\alpha_{\mathrm p}\LPIPS(\hat{\vx}_{\mathrm s},\hat{\vx}_{\mathrm t}).
\label{eq:l-s2t}
\end{equation}
Direct ground-truth supervision anchors the one-step output to the reference:
\begin{equation}
\Ls_{\mathrm{s}\rightarrow\mathrm{x}}=\beta_1\normone{\hat{\vx}_{\mathrm s}-\vx}
+\beta_{\mathrm p}\LPIPS(\hat{\vx}_{\mathrm s},\vx).
\label{eq:l-s2x}
\end{equation}
The two endpoint objectives are complemented by a ground-truth-to-student path that constrains the velocity field. For $\tau\sim\gU[\tau_{\min},\tau_{\max}]$, the inverse-flow source is
\begin{equation}
\hat{\vc}=\vx-\tau\vv_{\vphi}(\vx,\tau),
\label{eq:cinv}
\end{equation}
and maps it back to the high-quality endpoint by
\begin{equation}
\hat{\vx}_{\tau}=\hat{\vc}+\vv_{\vphi}(\hat{\vc},0).
\label{eq:xtau}
\end{equation}
The corresponding inverse-consistency objective is
\begin{equation}
\Ls_{\mathrm{x}\rightarrow\mathrm{s}}=\gamma_1\normone{\hat{\vx}_{\tau}-\vx}
+\gamma_{\mathrm p}\LPIPS(\hat{\vx}_{\tau},\vx).
\label{eq:l-x2s}
\end{equation}

\noindent\textbf{Training objective.} The student objective combines flow matching, distillation, and adversarial training:
\begin{align}
\Ls_{\mathrm{student}}{=}{}\lambda_{\mathrm v}\Ls_{\mathrm v}^{\mathrm s}
{+}\Ls_{\mathrm{s}\rightarrow\mathrm{t}}{+}\Ls_{\mathrm{s}\rightarrow\mathrm{x}}
{+}\Ls_{\mathrm{x}\rightarrow\mathrm{s}}{+}\lambda_{\mathrm a}\Ls_{\mathrm{adv}}.
\label{eq:lstudent}
\end{align}
With both teacher flows frozen, $\gS_{\vphi}$ performs complete LR-to-HR restoration in one pixel-space evaluation.

\begin{table*}[!t]\centering\footnotesize
\setlength{\tabcolsep}{3.6pt}
\begin{tabular}{lcccccccc}
\toprule
Method & PSNR$\uparrow$ & SSIM$\uparrow$ & MANIQA$\uparrow$ & MUSIQ$\uparrow$ & CLIP-IQA$\uparrow$ & LPIPS$\downarrow$ & DISTS$\downarrow$ & NIQE$\downarrow$ \\
\midrule
\resultgroup{Multi-step / flexible-step methods}
StableSR~\cite{wang2024stablesr} & 24.70 & 0.709 & 0.622 & 65.78 & 0.618 & 0.302 & 0.229 & 5.912 \\
DiffBIR~\cite{lin2024diffbir} & 24.75 & 0.657 & 0.625 & 64.98 & 0.646 & 0.364 & 0.231 & 5.535 \\
SeeSR~\cite{wu2024seesr} & 25.18 & 0.722 & 0.644 & \underline{69.77} & 0.661 & 0.301 & \underline{0.222} & 5.408 \\
PASD~\cite{yang2024pasd} & 25.21 & 0.680 & \underline{0.649} & 68.75 & 0.662 & 0.338 & 0.226 & 5.414 \\
ResShift~\cite{yue2023resshift} & \textbf{26.31} & \underline{0.742} & 0.528 & 58.43 & 0.544 & 0.346 & 0.250 & 7.263 \\
SUPIR~\cite{yu2024supir} & 23.65 & 0.662 & 0.578 & 62.09 & 0.671 & 0.354 & 0.249 & 6.110 \\
DreamClear~\cite{ai2024dreamclear} & 22.56 & 0.655 & 0.538 & 65.21 & 0.690 & 0.368 & 0.235 & 5.738 \\
InvSR~\cite{yue2025invsr} & 24.50 & 0.726 & 0.446 & 69.67 & \underline{0.692} & \underline{0.298} & 0.249 & \underline{5.219} \\
LinearSR~\cite{li2026linearsr} & 23.84 & 0.685 & 0.611 & 69.39 & 0.673 & 0.313 & 0.293 & 5.851 \\
\midrule
\textbf{Ours} & \underline{25.97} & \textbf{0.751} & 0.600 & 65.15 & 0.651 & \textbf{0.248} & \textbf{0.218} & \textbf{4.996} \\
\textbf{Ours-L} & 23.52 & 0.614 & \textbf{0.660} & \textbf{72.33} & \textbf{0.714} & 0.338 & 0.254 & 5.388 \\
\midrule
\resultgroup{Efficient / one-step methods}
SinSR~\cite{wang2024sinsr} & \underline{25.98} & 0.735 & 0.538 & 60.80 & 0.612 & 0.319 & 0.235 & 6.287 \\
OSEDiff~\cite{wu2024osediff} & 25.15 & 0.734 & 0.633 & 69.09 & 0.669 & 0.292 & 0.213 & 5.648 \\
S3Diff~\cite{zhang2024s3diff} & 25.03 & 0.732 & 0.626 & 67.89 & 0.672 & \underline{0.270} & \textbf{0.200} & 5.331 \\
AddSR~\cite{tai2026addsr} & 23.33 & 0.640 & \textbf{0.683} & \textbf{71.49} & 0.723 & 0.393 & 0.263 & 5.896 \\
D3SR~\cite{li2025d3sr} & 24.54 & 0.727 & 0.638 & 68.69 & 0.671 & 0.305 & 0.211 & \underline{5.096} \\
FLUX-SR~\cite{li2025fluxsr} & 24.83 & \underline{0.738} & \underline{0.651} & 70.08 & \underline{0.738} & 0.314 & 0.226 & 5.210 \\
AdcSR~\cite{chen2025adcsr} & 25.31 & 0.724 & 0.637 & 70.31 & 0.736 & 0.300 & 0.216 & 5.315 \\
TSD-SR~\cite{dong2025tsdsr} & 24.81 & 0.717 & 0.635 & 70.49 & 0.716 & 0.274 & \underline{0.210} & 5.130 \\
VOSR~\cite{wu2026vosr} &25.32 & 0.709 & 0.650 & 70.00 & 0.576 & 0.286 & \underline{0.210} & 5.239 \\
\midrule
\textbf{Ours} & \textbf{26.83} & \textbf{0.753} & 0.587 & 67.84 & 0.713 & \textbf{0.236} & 0.217 & 5.512 \\
\textbf{Ours-L} & 25.17 & 0.691 & 0.606 & \underline{71.43} & \textbf{0.759} & 0.298 & 0.269 & \textbf{4.941} \\
\bottomrule
\end{tabular}
\caption{Quantitative comparison on RealSR. Per-category best and second-best values are bolded and underlined.}
\label{tab:realsr}
\end{table*}

\section{Experiments}

\subsection{Setup}
\label{sec:setup}

\noindent\textbf{Datasets and degradation.} PixelIR is trained on high-quality image collections commonly used for Real-ISR~\cite{wang2021realesrgan,liang2021swinir}, using second-order Real-ESRGAN degradations~\cite{wang2021realesrgan} for synthetic pairs and aligned real pairs for domain adaptation. Evaluation covers three standard benchmarks~\cite{wu2024osediff,wang2024stablesr,wu2024seesr}: DIV2K-Val~\cite{agustsson2017div2k}, divided into 3000 non-overlapping $512{\times}512$ patches; RealSR~\cite{cai2019realsr}, containing 100 pairs; and DRealSR~\cite{wei2020drealsr}, containing 93 pairs. RealSR and DRealSR provide authentic LR--HR captures rather than synthetic degradations. All methods perform $4{\times}$ restoration from $128{\times}128$ to $512{\times}512$.

\noindent\textbf{Evaluation metrics.} We use PSNR, SSIM, MANIQA~\cite{yang2022maniqa}, MUSIQ~\cite{ke2021musiq}, CLIP-IQA~\cite{wang2023clipiqa}, LPIPS~\cite{zhang2018lpips}, DISTS~\cite{ding2022dists}, and NIQE~\cite{mittal2013niqe} to measure distortion fidelity and perceptual quality. Following the OSEDiff evaluation protocol~\cite{wu2024osediff}, PSNR and SSIM are computed on the Y channel without border cropping, while the remaining metrics are computed on RGB outputs. MANIQA, MUSIQ, CLIP-IQA, and NIQE are no-reference metrics, whereas LPIPS and DISTS measure reference-based perceptual distance. We compute all metrics with PyIQA~\cite{chen2022pyiqa}.

\noindent\textbf{Model variants.} We instantiate PixelIR with lightweight and prior-enhanced teacher--student branches. The lightweight branch pairs a 370.1M-parameter, eight-evaluation teacher with a 32.9M-parameter one-step student. PixelIR-L retains the same frozen fidelity flow but replaces the perceptual flow with the pretrained 1.3B PixelDiT prior adapted by rank-32 LoRA, while preserving the 4-step residual trajectory and residual composition. Its corresponding one-step student keeps the five resolutions and channel widths in \tabref{tab:student-arch}, increases the stage depths to $\{30,4,7,7,10\}$, and contains 132.8M parameters. The quantitative tables denote the lightweight and prior-enhanced branches as Ours and Ours-L, respectively; their teachers appear in the multi-step group and the corresponding distilled students appear in the one-step group.

\noindent\textbf{Compared methods.} We compare against the multi-step/flexible-step and efficient/one-step methods listed in the quantitative tables. Baseline values follow official reports under the shared three-benchmark protocol, and the best and second-best results are determined separately within each category. The DIV2K comparison and additional setup details are provided in the supplementary material.

\noindent\textbf{Implementation details.} All variants use AdamW and bf16 precision on one RTX PRO 6000 GPU. Both students follow the dual-anchor objective in \eqref{eq:lstudent} and require no text encoder at inference. Additional data mixtures, optimization settings, and text-conditioning details are provided in the supplementary material.
\vspace{-2mm}
\subsection{Main Results}
\label{sec:main-results}

\begin{table*}[!t]\centering\footnotesize
\setlength{\tabcolsep}{3.6pt}
\begin{tabular}{lcccccccc}
\toprule
Method & PSNR$\uparrow$ & SSIM$\uparrow$ & MANIQA$\uparrow$ & MUSIQ$\uparrow$ & CLIP-IQA$\uparrow$ & LPIPS$\downarrow$ & DISTS$\downarrow$ & NIQE$\downarrow$ \\
\midrule
\resultgroup{Multi-step / flexible-step methods}
StableSR~\cite{wang2024stablesr} & 28.03 & 0.754 & 0.560 & 58.51 & 0.636 & 0.328 & \underline{0.227} & 6.524 \\
DiffBIR~\cite{lin2024diffbir} & 26.71 & 0.657 & 0.593 & 61.07 & 0.639 & 0.456 & 0.275 & 6.312 \\
SeeSR~\cite{wu2024seesr} & 28.17 & 0.769 & 0.604 & 64.93 & 0.680 & 0.319 & 0.232 & 6.397 \\
PASD~\cite{yang2024pasd} & 27.36 & 0.707 & \underline{0.617} & 64.87 & 0.681 & 0.376 & 0.253 & \underline{5.547} \\
ResShift~\cite{yue2023resshift} & \underline{28.46} & 0.767 & 0.459 & 50.60 & 0.534 & 0.401 & 0.266 & 8.125 \\
SUPIR~\cite{yu2024supir} & 25.09 & 0.646 & 0.547 & 58.79 & 0.675 & 0.424 & 0.280 & 7.392 \\
DreamClear~\cite{ai2024dreamclear} & 24.48 & 0.651 & 0.447 & 65.83 & 0.662 & 0.397 & 0.244 & \textbf{5.133} \\
InvSR~\cite{yue2025invsr} & 27.63 & \textbf{0.796} & 0.461 & 67.46 & 0.692 & \underline{0.290} & 0.237 & 6.322 \\
LinearSR~\cite{li2026linearsr} & 26.91 & 0.719 & 0.581 & \underline{69.22} & \underline{0.713} & 0.358 & 0.300 & 6.965 \\
\midrule
\textbf{Ours} & \textbf{29.09} & \underline{0.789} & 0.602 & 60.52 & 0.668 & \textbf{0.269} & \textbf{0.219} & 5.927 \\
\textbf{Ours-L} & 26.80 & 0.690 & \textbf{0.642} & \textbf{71.23} & \textbf{0.729} & 0.344 & 0.271 & 5.976 \\
\midrule
\resultgroup{Efficient / one-step methods}
SinSR~\cite{wang2024sinsr} & 28.36 & 0.751 & 0.488 & 55.33 & 0.638 & 0.366 & 0.248 & 6.991 \\
OSEDiff~\cite{wu2024osediff} & 27.92 & 0.783 & 0.590 & 64.65 & 0.696 & 0.297 & 0.216 & 6.490 \\
S3Diff~\cite{zhang2024s3diff} & 27.39 & 0.747 & 0.572 & 64.16 & 0.716 & 0.313 & \textbf{0.211} & 6.170 \\
AddSR~\cite{tai2026addsr} & 26.72 & 0.712 & \textbf{0.626} & 66.33 & 0.723 & 0.398 & 0.271 & 7.669 \\
D3SR~\cite{li2025d3sr} & 26.98 & 0.714 & 0.596 & 67.28 & 0.709 & 0.308 & 0.223 & \textbf{5.523} \\
FLUX-SR~\cite{li2025fluxsr} & 27.29 & \underline{0.796} & 0.599 & \underline{68.79} & 0.673 & \underline{0.290} & 0.229 & 5.930 \\
AdcSR~\cite{chen2025adcsr} & 28.10 & 0.773 & 0.605 & 66.26 & 0.705 & 0.305 & 0.220 & 6.450 \\
TSD-SR~\cite{dong2025tsdsr} & 27.77 & 0.756 & 0.587 & 66.62 & \underline{0.734} & 0.297 & \underline{0.214} & 5.913 \\
VOSR~\cite{wu2026vosr} &27.66 & 0.732 & \underline{0.609} & 65.80 & 0.591 & 0.348 & 0.236 & \underline{5.773} \\
\midrule
\textbf{Ours} & \textbf{29.85} & \textbf{0.803} & 0.591 & 63.00 & 0.718 & \textbf{0.255} & 0.229 & 6.278 \\
\textbf{Ours-L} & \underline{28.39} & 0.750 & \underline{0.609} & \textbf{69.06} & \textbf{0.747} & 0.303 & 0.271 & 5.800 \\
\bottomrule
\end{tabular}
\caption{Quantitative comparison on DRealSR. Per-category best and second-best values are bolded and underlined.}
\label{tab:drealsr}
\end{table*}

\begin{figure*}[!t]\centering
\includegraphics[width=0.85\textwidth]{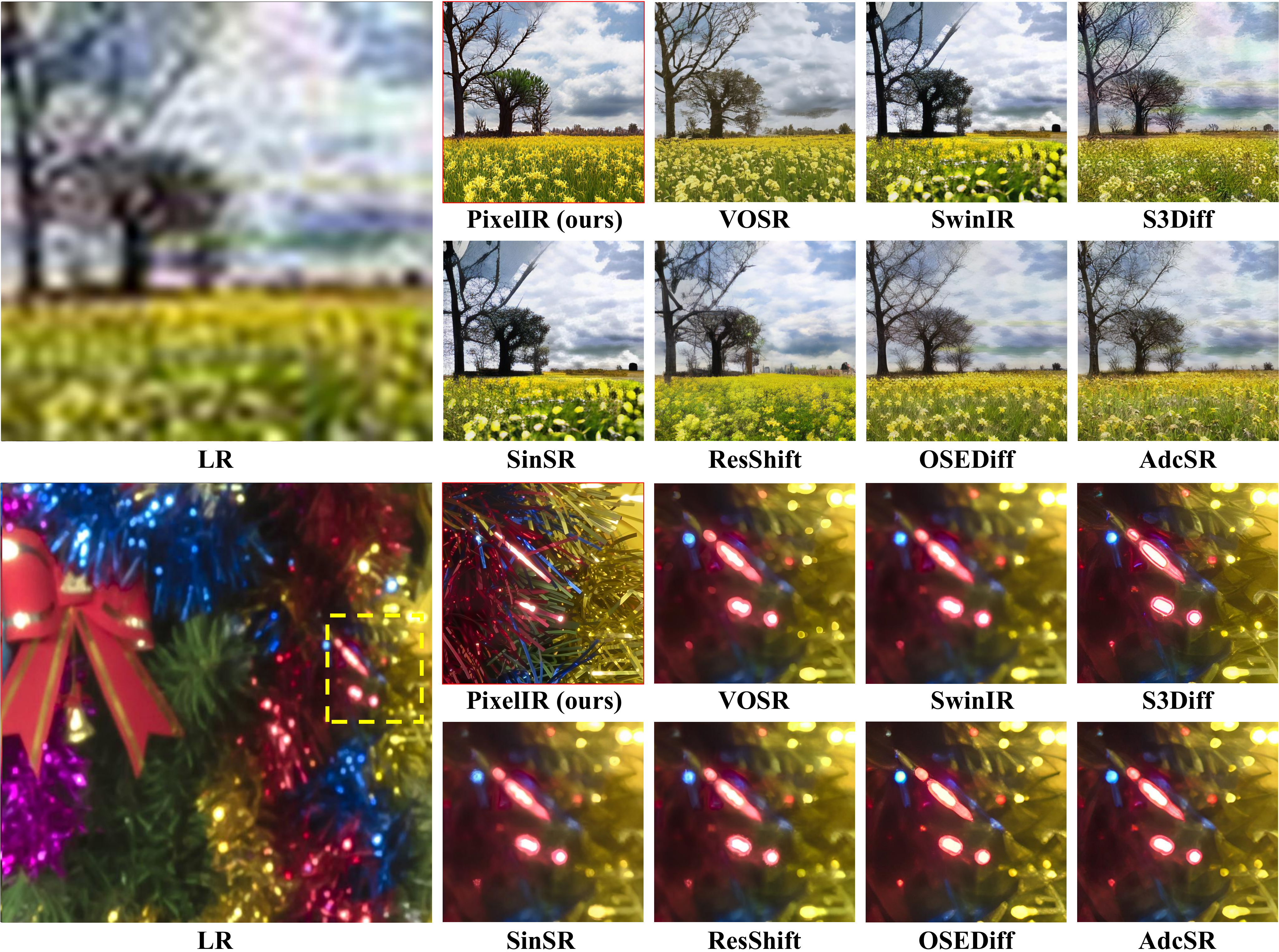}
\vspace{-2mm}
\caption{Qualitative comparison on challenging examples from LSDIR and RealSR.}
\vspace{-6mm}
\label{fig:qualitative}
\end{figure*}

\noindent\textbf{RealSR and DRealSR.} As shown in \twotabref{tab:realsr}{tab:drealsr}, the compact PixelIR student performs more favorably on real-world data. It achieves the best PSNR, SSIM, and LPIPS on both benchmarks. This indicates that the one-step student preserves strong structural fidelity and reference-based perceptual alignment under authentic degradations. Against recent one-step baselines, PixelIR better balances fidelity and reference-based perceptual quality on real data.

\noindent\textbf{DIV2K-Val.} The complete comparison is provided in the supplementary material. The compact student achieves a CLIP-IQA of 0.761, surpassing all prior methods. Taken together, the three datasets show that PixelIR balances reference fidelity, perceptual quality, and deployment efficiency rather than optimizing a single no-reference score.
\vspace{-1mm}
\subsection{Inference Efficiency}
\label{sec:efficiency}

\Tabref{tab:efficiency} compares the sampling steps, active parameters, MACs, and latency required for one $128{\to}512$ restoration. StableSR through AdcSR follow the unified third-party benchmark of AdcSR~\cite{chen2025adcsr}, while SANA-SR follows its official report. For ResShift and SinSR, we include both the 119M diffusion U-Net and the 55.3M VQGAN autoencoder executed at inference, giving 174.7M active parameters. On an RTX PRO 6000, we measure ours using exact state-dict parameters,  MACs, and CUDA-event latency over five bf16 batch-1 runs.

\begin{table}[!tbp]\centering\small
\setlength{\tabcolsep}{3.2pt}
\begin{tabular*}{\columnwidth}{@{\extracolsep{\fill}}lcccc@{}}
\toprule
Method & Steps & Params & MACs & Latency \\
\midrule
StableSR & 200 & 1410M & 79940G & 11.5\,s \\
DiffBIR  & 50  & 1717M & 24234G & 2.72\,s \\
SeeSR    & 50  & 2524M & 65857G & 4.30\,s \\
PASD     & 20  & 1900M & 29125G & 2.80\,s \\
ResShift & 15  & \underline{174.7M} & 5999G  & 0.71\,s \\
SinSR    & 1   & \underline{174.7M} & 3157G  & 0.13\,s \\
OSEDiff  & 1   & 1775M & 2265G & 0.11\,s \\
S3Diff   & 1   & 1327M & 2627G & 0.28\,s \\
AdcSR    & 1   & 456M & 496G & 0.030\,s \\
SANA-SR  & 1   & 344M  & \underline{407.95G} & \underline{0.019\,s} \\
\textbf{Ours} & 1 & \textbf{32.9M} & \textbf{89.7G} & \textbf{0.0085\,s} \\
\bottomrule
\end{tabular*}
\caption{Inference cost for a single $4\times$ SR image ($128{\to}512$). \textbf{Bold}/\underline{underline}: best/second-best per column; ties in step count are left unmarked.}
\vspace{-6mm}
\label{tab:efficiency}
\end{table}

\noindent\textbf{Efficiency.} \Tabref{tab:efficiency} shows that the compact PixelIR student further reduces end-to-end latency and MACs to 8.5ms and 89.7G, respectively. It performs the complete LR-to-HR mapping with one 32.9M-parameter pixel-space network and requires no VAE or text encoder at inference. Compared with the next-smallest ResShift and SinSR stacks, our method uses $5.3\times$ fewer parameters; compared with SANA-SR and OSEDiff, it uses $4.6\times$ and $25\times$ fewer MACs, respectively. Since published latency values use different hardware, parameter and MAC comparisons provide the more hardware-independent evidence. The left panel of \figref{fig:teaser} visualizes this regime.
\vspace{-1mm}
\subsection{Ablation Studies}
\label{sec:ablation}

We evaluate decoupling in the lightweight teacher and the direct endpoint losses of the compact student. Student-capacity, loss, and discriminator analyses are provided in the supplementary material.

\begin{figure}[!tbp]
\centering
\includegraphics[width=\columnwidth]{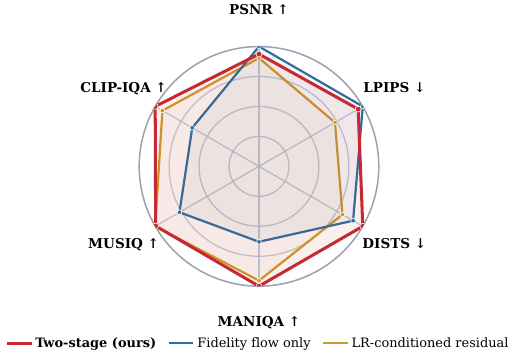}
\caption{Six-metric component profile on DIV2K. Each metric is direction-aligned and normalized by its best value; farther is better.}
\vspace{-3mm}
\label{fig:decouple}
\end{figure}

\noindent\textbf{Decoupling fidelity from perception.}
\Figref{fig:decouple} isolates the contributions of the fidelity and residual flows. Fidelity flow only removes the generative residual stage, while LR-conditioned residual serves as the matched single-stage baseline by removing the learned fidelity stage and conditioning residual generation directly on bicubic LR.

The fidelity flow alone favors distortion metrics but lacks generative perception, reaching only 51.33 MUSIQ and 0.474 CLIP-IQA. Residual generation raises them to 66.72 and 0.737 while improving DISTS from 0.213 to 0.193. Replacing $\hat{\vx}_{\mathrm b}$ with the raw LR condition further increases LPIPS from 0.234 to 0.305 and DISTS from 0.193 to 0.240, confirming that the frozen fidelity estimate provides both the output base and an informative detail condition.

\begin{table}[!tbp]\centering\small
\begin{tabular*}{\columnwidth}{@{\extracolsep{\fill}}lccc@{}}
\toprule
Configuration & PSNR & LPIPS$\downarrow$ & DISTS$\downarrow$ \\
\midrule
Full & 25.47 & \textbf{0.263} & \textbf{0.216} \\
$-\,\Ls_{\mathrm{s}\rightarrow\mathrm{x}}$ & 25.44 & 0.267 & 0.218 \\
$-\,\Ls_{\mathrm{s}\rightarrow\mathrm{t}}$ & 25.49 & 0.277 & 0.227 \\
$-\,\Ls_{\mathrm{s}\rightarrow\mathrm{t}}-\Ls_{\mathrm{s}\rightarrow\mathrm{x}}$ & 25.01 & 0.358 & 0.268 \\
\bottomrule
\end{tabular*}
\caption{Distillation objective ablation on DIV2K. Each row removes term(s) from \eqref{eq:lstudent}, all else fixed.}
\label{tab:distill-ablation}
\vspace{-6mm}
\end{table}

\noindent\textbf{Distillation objective components.}
\Tabref{tab:distill-ablation} evaluates the teacher and ground-truth anchors in \eqref{eq:lstudent} using LPIPS and DISTS.
Removing either direct endpoint loss causes only a minor change because the other still constrains the output. Removing both direct losses increases LPIPS from 0.263 to 0.358 and DISTS from 0.216 to 0.268, while MUSIQ rises from 68.9 to 72.2. Although inverse consistency, flow matching, and adversarial supervision remain, weaker endpoint constraints can improve a no-reference score through unmatched texture, confirming that the direct teacher and ground-truth losses preserve faithfulness.

\section{Conclusion}

We presented PixelIR, an efficient decouple-then-distill framework for Real-ISR. Motivated by the observation that existing generative restorers optimize faithful reconstruction and perceptual detail synthesis within a shared trajectory, we revisited Real-ISR from the perspectives of fidelity--perception decoupling and efficient one-step deployment. To this end, PixelIR combines a 4-step image flow for faithful reconstruction with a 4-step residual flow for complementary perceptual details. The resulting teacher is distilled into a progressively narrowed coarse-to-fine student under complementary teacher and ground-truth anchors. As a result, PixelIR establishes a stronger fidelity--perception operating point before transferring the complete restoration mapping to a single pixel-space evaluation. Extensive experiments show leading PSNR, SSIM, and LPIPS on RealSR and DRealSR. With 32.9M parameters and 89.7G MACs, the compact student runs in 8.5ms on an RTX PRO 6000, demonstrating a favorable fidelity--perception--efficiency balance.

\begin{small}
\bibliography{main}
\end{small}


\end{document}